\documentclass[10pt,twocolumn,letterpaper]{article}

\usepackage{cvpr}              
\usepackage{url}            
\usepackage{booktabs}       
\usepackage{amsfonts}       
\usepackage{nicefrac}       
\usepackage{microtype}      
\usepackage{times}
\usepackage{epsfig}
\usepackage{graphicx}
\usepackage{caption}
\usepackage{amsmath}
\usepackage{amssymb}
\usepackage{multirow}
\usepackage{color}
\usepackage{bbding}
\usepackage{makecell}
\usepackage{import}
\usepackage{tabularx}
\usepackage{array}
\usepackage{dsfont}
\usepackage{bbm}
\renewcommand{\thefootnote}{}
\usepackage[dvipsnames, table]{xcolor}

\usepackage[pagebackref,breaklinks,colorlinks,citecolor=cvprblue]{hyperref}
\usepackage[misc]{ifsym}
\definecolor{cvprblue}{rgb}{0.21,0.49,0.74}
\definecolor{aliceblue}{rgb}{0.94, 0.97, 1.0}
\makeatletter
\newcommand{\thickhline}{%
 \noalign {\ifnum 0=`}\fi \hrule height 1pt
 \futurelet \reserved@a \@xhline
}

\DeclareMathAlphabet\mathbfcal{OMS}{cmsy}{b}{n}

\def\paperID{4117} 
\def\confName{CVPR}
\def\confYear{2024}

\begin{document}

\title{DETRs Beat YOLOs on Real-time Object Detection}
\author{Yian Zhao\textsuperscript{1,2}$^\dag$ \quad Wenyu Lv\textsuperscript{1}$^{\dag\ddag}$ \quad Shangliang Xu\textsuperscript{1} \quad Jinman Wei\textsuperscript{1} \quad Guanzhong Wang\textsuperscript{1} \quad \\
\vspace*{1mm}
Qingqing Dang\textsuperscript{1} \quad Yi Liu\textsuperscript{1} \quad Jie Chen\textsuperscript{2\ \Letter} \\
\small \textsuperscript{1}Baidu Inc, Beijing, China \quad
\small \textsuperscript{2}School of Electronic and Computer Engineering, Peking University, Shenzhen, China \\
\small \href{mailto:zhaoyian@stu.pku.edu.cn}{zhaoyian@stu.pku.edu.cn} \quad \small \href{mailto:lvwenyu01@baidu.com}{lvwenyu01@baidu.com}  \quad \small \href{mailto:jiechen2019@pku.edu.cn}{jiechen2019@pku.edu.cn}
\vspace*{-7mm}
}
\maketitle
\begin{abstract}
The YOLO series has become the most popular framework for real-time object detection due to its reasonable trade-off between speed and accuracy. 
However, we observe that the speed and accuracy of YOLOs are negatively affected by the NMS. 
Recently, end-to-end Transformer-based detectors~(DETRs) have provided an alternative to eliminating NMS. 
Nevertheless, the high computational cost limits their practicality and hinders them from fully exploiting the advantage of excluding NMS. 
In this paper, we propose the \textbf{R}eal-\textbf{T}ime \textbf{DE}tection \textbf{TR}ansformer~(RT-DETR), the first real-time end-to-end object detector to our best knowledge that addresses the above dilemma. 
We build RT-DETR in two steps, drawing on the advanced DETR: first we focus on maintaining accuracy while improving speed, followed by maintaining speed while improving accuracy.
Specifically, we design an efficient hybrid encoder to expeditiously process multi-scale features by decoupling intra-scale interaction and cross-scale fusion to improve speed. 
Then, we propose the uncertainty-minimal query selection to provide high-quality initial queries to the decoder, thereby improving accuracy.
In addition, RT-DETR supports flexible speed tuning by adjusting the number of decoder layers to adapt to various scenarios without retraining.
Our RT-DETR-R50 / R101 achieves $53.1\%$ / $54.3\%$ AP on COCO and $108$ / $74$ FPS on T4 GPU, outperforming previously advanced YOLOs in both speed and accuracy.
%
%
Furthermore, RT-DETR-R50 outperforms DINO-R50 by $2.2\%$ AP in accuracy and about $21$ times in FPS.
After pre-training with Objects365, RT-DETR-R50 / R101 achieves $55.3\%$ / $56.2\%$ AP.
%
The project page: \href{https://zhao-yian.github.io/RTDETR/}{https://zhao-yian.github.io/RTDETR}.
\end{abstract}
\footnote{\Letter\ Corresponding author. $^\dag$Equal contribution. $^\ddag$ Project leader.}
\vspace*{-4mm}
\section{Introduction}
\label{sec:intro}
\begin{figure}[!ht]
\centering
\includegraphics[width=\linewidth]{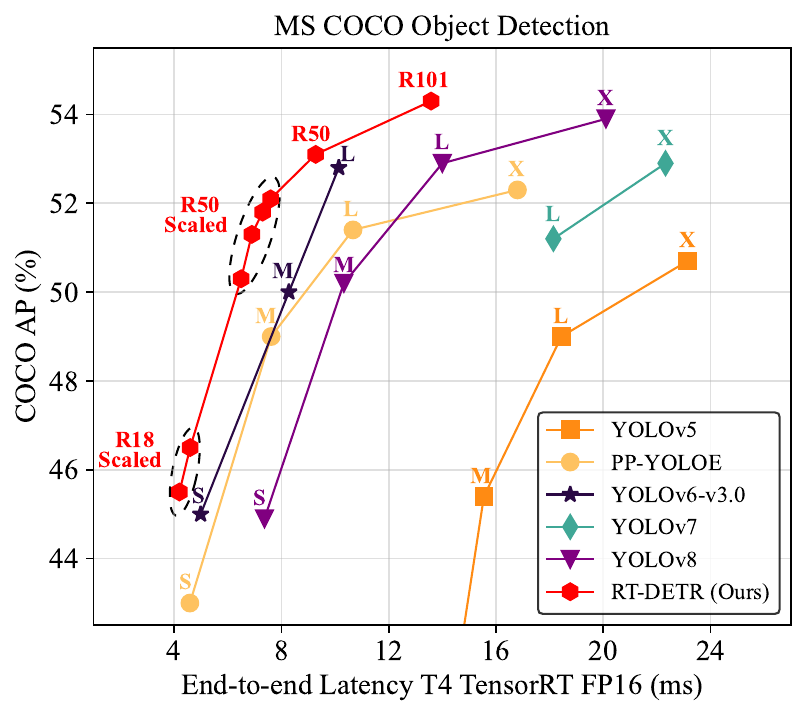} 
\vspace*{-8mm}
\caption{Compared to previously advanced real-time object detectors, our RT-DETR achieves state-of-the-art performance.}
\vspace*{-6mm}
\label{fig:yolo series}
\end{figure}

Real-time object detection is an important area of research and has a wide range of applications, such as object tracking~\cite{zeng2022motr}, video surveillance~\cite{nawaratne2019spatiotemporal}, and autonomous driving~\cite{bogdoll2022anomaly}, \textit{etc.}
Existing real-time detectors generally adopt the CNN-based architecture, the most famous of which is the YOLO detectors~\cite{redmon2018yolov3,bochkovskiy2020yolov4,yolov5v7.0,long2020pp,huang2021pp,xu2022pp,li2023yolov6v3,ge2021yolox,wang2023yolov7,yolov8} due to their reasonable trade-off between speed and accuracy.
However, these detectors typically require Non-Maximum Suppression~(NMS) for post-processing, which not only slows down the inference speed but also introduces hyperparameters that cause instability in both the speed and accuracy.
Moreover, considering that different scenarios place different emphasis on recall and accuracy, it is necessary to carefully select the appropriate NMS thresholds, which hinders the development of real-time detectors.

Recently, the end-to-end Transformer-based detectors~(DETRs)~\cite{carion2020end,sun2021sparse,zhu2020deformable,meng2021conditional,wang2022anchor,liu2021dab,li2022dn,zhang2022dino} have received extensive attention from the academia due to their streamlined architecture and elimination of hand-crafted components.
%
%
However, their high computational cost prevents them from meeting real-time detection requirements, so the NMS-free architecture does not demonstrate an inference speed advantage. 
%
%
%
%
This inspires us to explore whether DETRs can be extended to real-time scenarios and outperform the advanced YOLO detectors in both speed and accuracy, eliminating the delay caused by NMS for real-time object detection.

To achieve the above goal, we rethink DETRs and conduct detailed analysis of key components to reduce unnecessary computational redundancy and further improve accuracy.
For the former, we observe that although the introduction of multi-scale features is beneficial in accelerating the training convergence~\cite{zhu2020deformable}, it leads to a significant increase in the length of the sequence feed into the encoder.
The high computational cost caused by the interaction of multi-scale features makes the Transformer encoder the computational bottleneck.
Therefore, implementing the real-time DETR requires a redesign of the encoder.
And for the latter, previous works~\cite{zhu2020deformable,yao2021efficient,zhang2022dino} show that the hard-to-optimize object queries hinder the performance of DETRs and propose the query selection schemes to replace the vanilla learnable embeddings with encoder features.
However, we observe that the current query selection directly adopt classification scores for selection, ignoring the fact that the detector are required to simultaneously model the category and location of objects, both of which determine the quality of the features.
This inevitably results in encoder features with low localization confidence being selected as initial queries, thus leading to a considerable level of uncertainty and hurting the performance of DETRs.
We view query initialization as a breakthrough to further improve performance.
%

%
In this paper, we propose the \textbf{R}eal-\textbf{T}ime \textbf{DE}tection \textbf{TR}ansformer~(RT-DETR), the first real-time end-to-end object detector to our best knowledge.
To expeditiously process multi-scale features, we design an efficient hybrid encoder to replace the vanilla Transformer encoder, which significantly improves inference speed by decoupling the intra-scale interaction and cross-scale fusion of features with different scales.
To avoid encoder features with low localization confidence being selected as object queries, we propose the uncertainty-minimal query selection, which provides high-quality initial queries to the decoder by explicitly optimizing the uncertainty, thereby increasing the accuracy.
Furthermore, RT-DETR supports flexible speed tuning to accommodate various real-time scenarios without retraining, thanks to the multi-layer decoder architecture of DETR.

RT-DETR achieves an ideal trade-off between the speed and accuracy.
Specifically, RT-DETR-R50 achieves $53.1\%$ AP on COCO \texttt{val2017} and $108$ FPS on T4 GPU, while RT-DETR-R101 achieves $54.3\%$ AP and $74$ FPS, outperforming $L$ and $X$ models of previously advanced YOLO detectors in both speed and accuracy, \Cref{fig:yolo series}.
We also develop scaled RT-DETRs by scaling the encoder and decoder with smaller backbones, which outperform the lighter YOLO detectors~($S$ and $M$ models).
Furthermore, RT-DETR-R50 outperforms DINO-Deformable-DETR-R50 by $2.2\%$ AP~($53.1\%$ AP vs $50.9\%$ AP) in accuracy and by about $21$ times in FPS~($108$ FPS vs $5$ FPS), significantly improves accuracy and speed of DETRs.
After pre-training with Objects365~\cite{shao2019objects365}, RT-DETR-R50 / R101 achieves $55.3\%$ / $56.2\%$ AP, resulting in surprising performance improvements. 
More experimental results are provided in the Appendix.

The main contributions are summarized as: 
(\romannumeral1). We propose the first real-time end-to-end object detector called RT-DETR, which not only outperforms the previously advanced YOLO detectors in both speed and accuracy but also eliminates the negative impact caused by NMS post-processing on real-time object detection; 
(\romannumeral2). We quantitatively analyze the impact of NMS on the speed and accuracy of YOLO detectors, and establish an end-to-end speed benchmark to test the end-to-end inference speed of real-time detectors; 
(\romannumeral3). The proposed RT-DETR supports flexible speed tuning by adjusting the number of decoder layers to accommodate various scenarios without retraining.
\section{Related Work}

\subsection{Real-time Object Detectors}
YOLOv1~\cite{redmon2016you} is the first CNN-based one-stage object detector to achieve true real-time object detection.
Through years of continuous development, the YOLO detectors have outperformed other one-stage object detectors~\cite{liu2016ssd,lin2017focal} and become the synonymous with the real-time object detector.
YOLO detectors can be classified into two categories: anchor-based~\cite{redmon2017yolo9000,redmon2018yolov3,bochkovskiy2020yolov4,long2020pp,huang2021pp,wang2021scaled,yolov5v7.0,wang2023yolov7} and anchor-free~\cite{ge2021yolox,xu2022pp,li2023yolov6v3,yolov8}, which achieve a reasonable trade-off between speed and accuracy and are widely used in various practical scenarios.
%
%
These advanced real-time detectors produce numerous overlapping boxes and require NMS post-processing, which slows down their speed.
%

\subsection{End-to-end Object Detectors}
%
End-to-end object detectors are well-known for their streamlined pipelines.
Carion \etal~\cite{carion2020end} first propose the end-to-end detector based on Transformer called DETR, which has attracted extensive attention due to its distinctive features. 
Particularly, DETR eliminates the hand-crafted anchor and NMS components.
Instead, it employs bipartite matching and directly predicts the one-to-one object set.
%
%
Despite its obvious advantages, DETR suffers from several problems: slow training convergence, high computational cost, and hard-to-optimize queries. 
Many DETR variants have been proposed to address these issues.
\noindent \textbf{Accelerating convergence.}
Deformable-DETR~\cite{zhu2020deformable} accelerates training convergence with multi-scale features by enhancing the efficiency of the attention mechanism.
DAB-DETR~\cite{liu2021dab} and DN-DETR~\cite{li2022dn} further improve performance by introducing the iterative refinement scheme and denoising training.
Group-DETR~\cite{chen2022group} introduces group-wise one-to-many assignment.
\noindent \textbf{Reducing computational cost.}
Efficient DETR~\cite{yao2021efficient} and Sparse DETR~\cite{roh2021sparse} reduce the computational cost by reducing the number of encoder and decoder layers or the number of updated queries.
Lite DETR~\cite{li2023lite} enhances the efficiency of encoder by reducing the update frequency of low-level features in an interleaved way.
%
%
\noindent \textbf{Optimizing query initialization.}
Conditional DETR~\cite{meng2021conditional} and Anchor DETR~\cite{wang2022anchor} decrease the optimization difficulty of the queries.
Zhu \etal~\cite{zhu2020deformable} propose the query selection for two-stage DETR, and DINO~\cite{zhang2022dino} suggests the mixed query selection to help better initialize queries. 
Current DETRs are still computationally intensive and are not designed to detect in real time.
Our RT-DETR vigorously explores computational cost reduction and attempts to optimize query initialization, outperforming state-of-the-art real-time detectors.
\begin{figure}[!t]
\centering
\includegraphics[width=\linewidth]{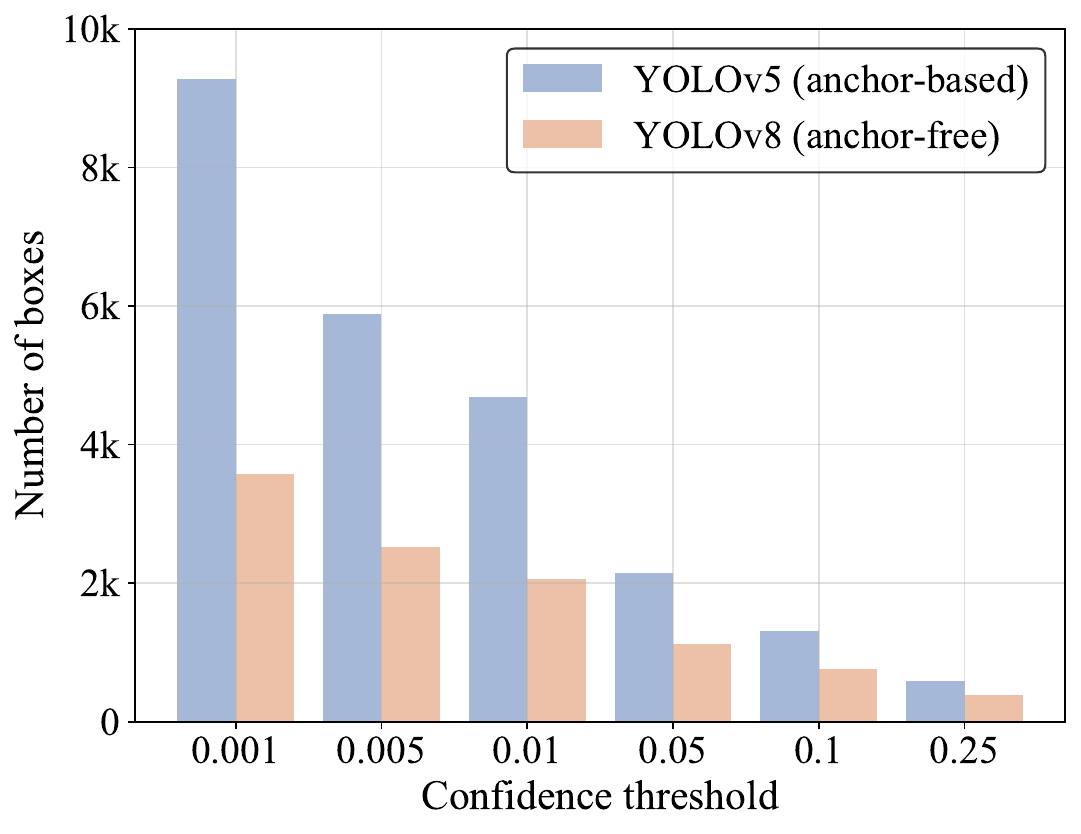}
\vspace*{-7mm}
\caption{The number of boxes at different confidence thresholds.}
\vspace*{-6mm}
\label{fig:num_boxes}
\end{figure}
\section{End-to-end Speed of Detectors}
\label{sec: speed}
\subsection{Analysis of NMS}
\label{subsec: analysis}
NMS is a widely used post-processing algorithm in object detection, employed to eliminate overlapping output boxes. 
Two thresholds are required in NMS: confidence threshold and IoU threshold.
Specifically, the boxes with scores below the confidence threshold are directly filtered out, and whenever the IoU of any two boxes exceeds the IoU threshold, the box with the lower score will be discarded.
This process is performed iteratively until all boxes of every category have been processed.
Thus, the execution time of NMS primarily depends on the number of boxes and two thresholds.
To verify this observation, we leverage YOLOv5~\cite{yolov5v7.0} (anchor-based) and YOLOv8~\cite{yolov8} (anchor-free) for analysis.

We first count the number of boxes remaining after filtering the output boxes with different confidence thresholds on the same input.
We sample values from $0.001$ to $0.25$ as confidence thresholds to count the number of remaining boxes of the two detectors and plot them on a bar graph, which intuitively reflects that NMS is sensitive to its hyperparameters, \Cref{fig:num_boxes}.
As the confidence threshold increases, more prediction boxes are filtered out, and the number of remaining boxes that need to calculate IoU decreases, thus reducing the execution time of NMS.

\begin{table}[!t]
\centering
\begin{minipage}{0.49\linewidth}
\centering
    \renewcommand{\arraystretch}{1.15}
    \setlength{\tabcolsep}{4.43pt}
    \begin{tabular}{c|cc}
    \toprule
    \textbf{\makecell[c]{IoU thr.\\\small(Conf=0.001)}}  &  \textbf{\makecell[c]{AP \\\small(\%)}} & \textbf{\makecell[c]{NMS\\\small(ms)}} \\ 
    \midrule
    0.5 & 52.1 & 2.24 \\
    \midrule
    0.6 & 52.6 & 2.29 \\
    \midrule
    0.8 & 52.8 & 2.46 \\
    \bottomrule
    \end{tabular}
\end{minipage}
\begin{minipage}{0.49\linewidth}
\centering
    \renewcommand{\arraystretch}{1.15}
    \setlength{\tabcolsep}{4.43pt}
    \begin{tabular}{c|cc}
    \toprule
    \textbf{\makecell[c]{Conf thr.\\\small(IoU=0.7)}}  &  \textbf{\makecell[c]{AP \\\small(\%)}} & \textbf{\makecell[c]{NMS\\\small(ms)}} \\
    \midrule
    0.001 & 52.9 & 2.36 \\
    \midrule
    0.01 & 52.4 & 1.73 \\
    \midrule
    0.05 & 51.2 & 1.06 \\
    \bottomrule
    \end{tabular}
\end{minipage}
\caption{The effect of IoU threshold and confidence threshold on  accuracy and NMS execution time.}
\vspace*{-6mm}
\label{tab:speed}
\end{table}

%
Furthermore, we use YOLOv8 to evaluate the accuracy on the COCO \texttt{val2017} and test the execution time of the NMS operation under different hyperparameters.
Note that the NMS operation we adopt refers to the TensorRT \texttt{efficientNMSPlugin}\footnote{\textcolor{magenta}{\scriptsize\url{https://github.com/NVIDIA/TensorRT/tree/release/8.6/plugin/efficientNMSPlugin}}}, which involves multiple kernels, including \texttt{EfficientNMSFilter}, \texttt{RadixSort}, \texttt{EfficientNMS}, \etc, and we only report the execution time of the \texttt{EfficientNMS} kernel.
We test the speed on T4 GPU with TensorRT FP16, and the input and preprocessing remain consistent.
The hyperparameters and the corresponding results are shown in ~\Cref{tab:speed}.
From the results, we can conclude that the execution time of the \texttt{EfficientNMS} kernel increases as the confidence threshold decreases or the IoU threshold increases.
The reason is that the high confidence threshold directly filters out more prediction boxes, whereas the high IoU threshold filters out fewer prediction boxes in each round of screening.
We also visualize the predictions of YOLOv8 with different NMS thresholds in Appendix. 
The results show that inappropriate confidence thresholds lead to significant false positives or false negatives by the detector.
With a confidence threshold of $0.001$ and an IoU threshold of $0.7$, YOLOv8 achieves the best AP results, but the corresponding NMS time is at a higher level.
Considering that YOLO detectors typically report the model speed and exclude the NMS time, thus an end-to-end speed benchmark needs to be established.

\begin{figure*}[!ht]
\centering
\includegraphics[width=\linewidth]{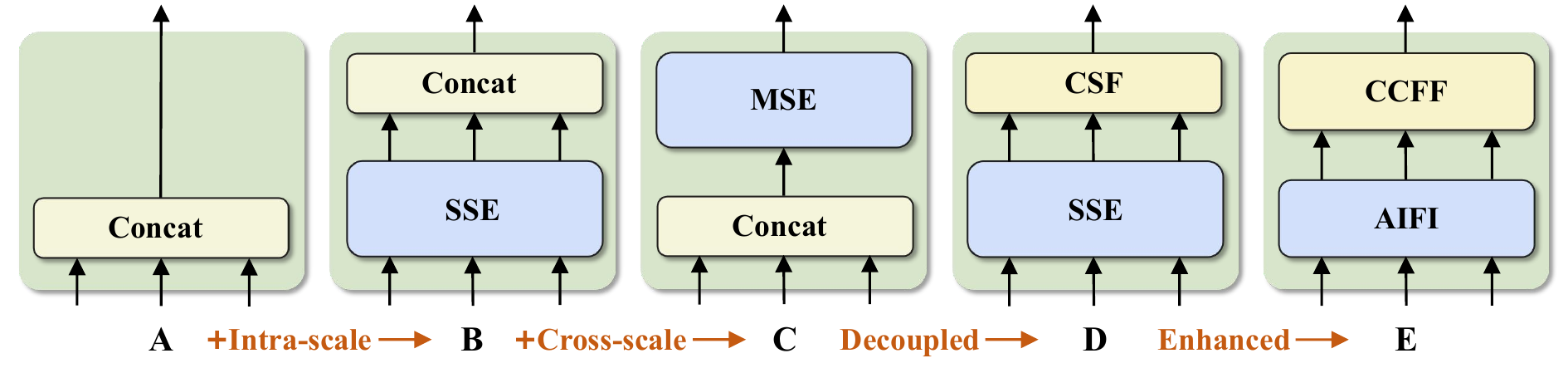} 
\vspace*{-6mm}
\caption{The encoder structure for each variant. \textbf{SSE} represents the single-scale Transformer encoder, \textbf{MSE} represents the multi-scale Transformer encoder, and \textbf{CSF} represents cross-scale fusion. \textbf{AIFI} and \textbf{CCFF} are the two modules designed into our hybrid encoder.}
\vspace*{-6mm}
\label{fig:variants}
\end{figure*}

\subsection{End-to-end Speed Benchmark}
\label{subsec: benchmark}
To enable a fair comparison of the end-to-end speed of various real-time detectors, we establish an end-to-end speed benchmark. 
Considering that the execution time of NMS is influenced by the input, it is necessary to choose a benchmark dataset and calculate the average execution time across multiple images.
We choose COCO \texttt{val2017}~\cite{lin2014microsoft} as the benchmark dataset and append the NMS post-processing plugin of TensorRT for YOLO detectors as mentioned above.
Specifically, we test the average inference time of the detector according to the NMS thresholds of the corresponding accuracy taken on the benchmark dataset, excluding \texttt{I/O} and \texttt{MemoryCopy} operations.
We utilize the benchmark to test the end-to-end speed of anchor-based detectors YOLOv5~\cite{yolov5v7.0} and YOLOv7~\cite{wang2023yolov7}, as well as anchor-free detectors PP-YOLOE~\cite{xu2022pp}, YOLOv6~\cite{li2023yolov6v3} and YOLOv8~\cite{yolov8} on T4 GPU with TensorRT FP16.
According to the results~(\textit{cf.~\Cref{tab:main}}), we conclude that \textit{anchor-free detectors outperform anchor-based detectors with equivalent accuracy for YOLO detectors because the former require less NMS time than the latter}.
The reason is that anchor-based detectors produce more prediction boxes than anchor-free detectors~(three times more in our tested detectors).

\section{The Real-time DETR}
\label{sec: method}
\subsection{Model Overview} 
\begin{figure*}[!ht]
    \centering
    \includegraphics[width=1.0\linewidth]{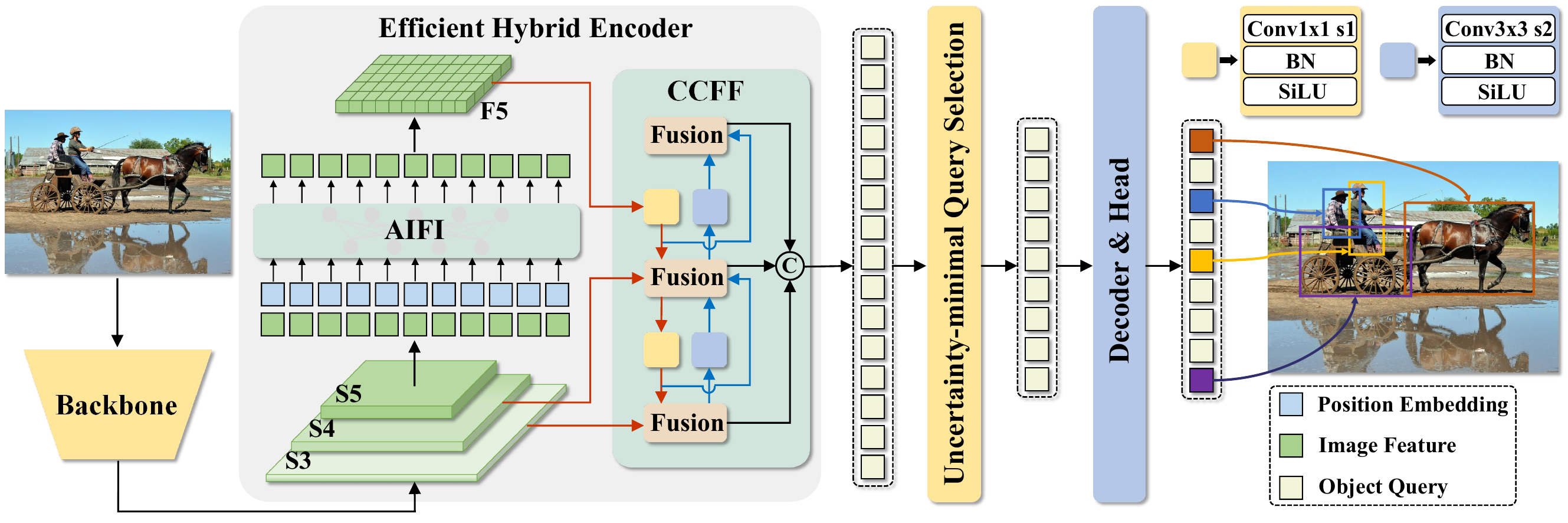}
    \vspace*{-6mm}
    \caption{
Overview of RT-DETR. We feed the features from the last three stages of the backbone into the encoder.
The efficient hybrid encoder transforms multi-scale features into a sequence of image features through the Attention-based Intra-scale Feature Interaction~(AIFI) and the CNN-based Cross-scale Feature Fusion~(CCFF).
Then, the uncertainty-minimal query selection selects a fixed number of encoder features to serve as initial object queries for the decoder.
Finally, the decoder with auxiliary prediction heads iteratively optimizes object queries to generate categories and boxes.
}
\vspace*{-6mm}
\label{fig:overview}
\end{figure*}

RT-DETR consists of a backbone, an efficient hybrid encoder, and a Transformer decoder with auxiliary prediction heads.
The overview of RT-DETR is illustrated in~\Cref{fig:overview}.
Specifically, we feed the features from the last three stages of the backbone $\{ \mathbfcal{S}_3, \mathbfcal{S}_4, \mathbfcal{S}_5 \}$ into the encoder.
The efficient hybrid encoder transforms multi-scale features into a sequence of image features through intra-scale feature interaction and cross-scale feature fusion~(\textit{cf.}~\cref{subsec:encoder}).
Subsequently, the uncertainty-minimal query selection is employed to select a fixed number of encoder features to serve as initial object queries for the decoder~(\textit{cf.}~\cref{subsec:query selection}).
Finally, the decoder with auxiliary prediction heads iteratively optimizes object queries to generate categories and boxes.

\subsection{Efficient Hybrid Encoder} 
\label{subsec:encoder}
\noindent \textbf{Computational bottleneck analysis.}
The introduction of multi-scale features accelerates training convergence and improves performance~\cite{zhu2020deformable}.
However, although the deformable attention reduces the computational cost, the sharply increased sequence length still causes the encoder to become the computational bottleneck.
As reported in Lin \etal~\cite{lin2022d}, the encoder accounts for $49\%$ of the GFLOPs but contributes only $11\%$ of the AP in Deformable-DETR.
To overcome this bottleneck, we first analyze the computational redundancy present in the multi-scale Transformer encoder.
Intuitively, high-level features that contain rich semantic information about objects are extracted from low-level features, making it redundant to perform feature interaction on the concatenated multi-scale features.
Therefore, we design a set of variants with different types of the encoder to prove that the simultaneous intra-scale and cross-scale feature interaction is inefficient, \Cref{fig:variants}.
%
%
Specially, we use DINO-Deformable-R50 with the smaller size data reader and lighter decoder used in RT-DETR for experiments and first remove the multi-scale Transformer encoder in DINO-Deformable-R50 as variant A.
Then, different types of the encoder are inserted to produce a series of variants based on A, elaborated as follows~(Detailed indicators of each variant are referred to in~\Cref{tab:encoder}):
\begin{itemize}
    \item A $\rightarrow$ B: Variant B inserts a single-scale Transformer encoder into A, which uses one layer of Transformer block. The multi-scale features share the encoder for intra-scale feature interaction and then concatenate as output.
    \item B $\rightarrow$ C: 
    Variant C introduces cross-scale feature fusion based on B and feeds the concatenated features into the multi-scale Transformer encoder to perform simultaneous intra-scale and cross-scale feature interaction.
    \item C $\rightarrow$ D: 
    Variant D decouples intra-scale interaction and cross-scale fusion by utilizing the single-scale Transformer encoder for the former and a PANet-style~\cite{liu2018path} structure for the latter.
    \item D $\rightarrow$ E: 
    Variant E enhances the intra-scale interaction and cross-scale fusion based on D, adopting an efficient hybrid encoder designed by us.
\end{itemize}

\begin{figure}[!ht]
    \centering
    \includegraphics[width=\linewidth]{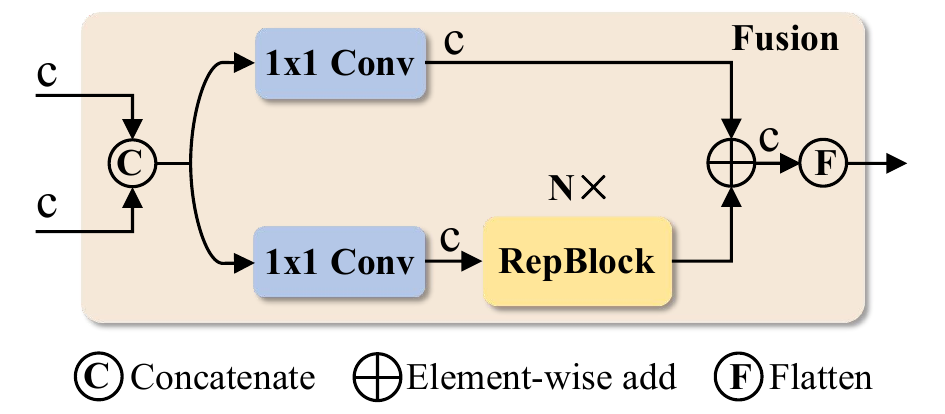}
    \vspace*{-7mm}
    \caption{The fusion block in CCFF.}
    \vspace*{-5mm}
 \label{fig:encoder_block}
\end{figure}

\noindent \textbf{Hybrid design.}
%
%
Based on the above analysis, we rethink the structure of the encoder and propose an \textit{efficient hybrid encoder}, consisting of two modules, namely the Attention-based Intra-scale Feature Interaction~(AIFI) and the CNN-based Cross-scale Feature Fusion~(CCFF).
Specifically, AIFI further reduces the computational cost based on variant D by performing the intra-scale interaction only on $\mathbfcal{S}_5$ with the single-scale Transformer encoder. 
The reason is that applying the self-attention operation to high-level features with richer semantic concepts captures the connection between conceptual entities, which facilitates the localization and recognition of objects by subsequent modules. 
However, the intra-scale interactions of lower-level features are unnecessary due to the lack of semantic concepts and the risk of duplication and confusion with high-level feature interactions.
To verify this opinion, we perform the intra-scale interaction only on $\mathbfcal{S}_5$ in variant D, and the experimental results are reported in~\Cref{tab:encoder}~(see row D$_{\mathbfcal{S}_5}$).
Compared to D, D$_{\mathbfcal{S}_5}$ not only significantly reduces latency~($35\%$ faster), but also improves accuracy~($0.4\%$ AP higher).
CCFF is optimized based on the cross-scale fusion module, which inserts several fusion blocks consisting of convolutional layers into the fusion path.
The role of the fusion block is to fuse two adjacent scale features into a new feature, and its structure is illustrated in~\Cref{fig:encoder_block}.
The fusion block contains two $1\times1$ convolutions to adjust the number of channels, $N$ \textit{RepBlock}s composed of RepConv~\cite{ding2021repvgg} are used for feature fusion, and the two-path outputs are fused by element-wise add.
We formulate the calculation of the hybrid encoder as:
\begin{equation}
\begin{split}
\label{equ:encoder}
    \mathbfcal{Q} & = \mathbfcal{K} = \mathbfcal{V} = \texttt{Flatten}(\mathbfcal{S}_5), \\
    \mathbfcal{F}_5&=\texttt{Reshape}(\texttt{AIFI}(\mathbfcal{Q},\mathbfcal{K},\mathbfcal{V})), \\
    \mathbfcal{O} & = \texttt{CCFF}(\{\mathbfcal{S}_3,\mathbfcal{S}_4,\mathbfcal{F}_5\}),
\end{split}
\end{equation}
where $\texttt{Reshape}$ represents restoring the shape of the flattened feature to the same shape as $\mathbfcal{S}_5$.
\subsection{Uncertainty-minimal Query Selection}
\label{subsec:query selection}
%
%
To reduce the difficulty of optimizing object queries in DETR, several subsequent works~\cite{zhu2020deformable,yao2021efficient,zhang2022dino} propose query selection schemes, which have in common that they use the confidence score to select the top $K$ features from the encoder to initialize object queries~(or just position queries).
The confidence score represents the likelihood that the feature includes foreground objects.
%
%
%
%
Nevertheless, the detector are required to simultaneously model the category and location of objects, both of which determine the quality of the features.
Hence, the performance score of the feature is a latent variable that is jointly correlated with both classification and localization.
Based on the analysis, the current query selection lead to a considerable level of uncertainty in the selected features, resulting in sub-optimal initialization for the decoder and hindering the performance of the detector.

To address this problem, we propose the uncertainty minimal query selection scheme, which explicitly constructs and optimizes the epistemic uncertainty to model the joint latent variable of encoder features, thereby providing high-quality queries for the decoder.
Specifically, the feature uncertainty $\mathcal{U}$ is defined as the discrepancy between the predicted distributions of localization $\mathcal{P}$ and classification $\mathcal{C}$ in~\cref{equ:uncertainty}.
%
%
%
To minimize the uncertainty of the queries, we integrate the uncertainty into the loss function for the gradient-based optimization in~\cref{equ:loss}.
\begin{equation}
\label{equ:uncertainty}
\mathcal{U}(\hat{\mathbfcal{X}}) = \| \mathcal{P}(\hat{\mathbfcal{X}}) - \mathcal{C}(\hat{\mathbfcal{X}}) \|, \hat{\mathbfcal{X}} \in \mathbb{R}^D
\end{equation}
\begin{equation}
\begin{split}
\label{equ:loss}
\mathcal{L}(\hat{\mathbfcal{X}}, \hat{\mathbfcal{Y}}, \mathbfcal{Y}) = \mathcal{L}_{box}(\hat{\mathbf{b}}, \mathbf{b}) + \mathcal{L}_{cls}(\mathcal{U}(\hat{\mathbfcal{X}}), \hat{\mathbf{c}}, \mathbf{c})
\end{split}
\end{equation}
%
where $\hat{\mathbfcal{Y}}$ and $\mathbfcal{Y}$ denote the prediction and ground truth, $\hat{\mathbfcal{Y}} = \{\hat{\mathbf{c}}, \hat{\mathbf{b}}\}$, $\hat{\mathbf{c}}$ and $\hat{\mathbf{b}}$ represent the category and bounding box respectively, $\hat{\mathbfcal{X}}$ represent the encoder feature.

\begin{figure}[t]
\centering
\includegraphics[width=0.95\linewidth]{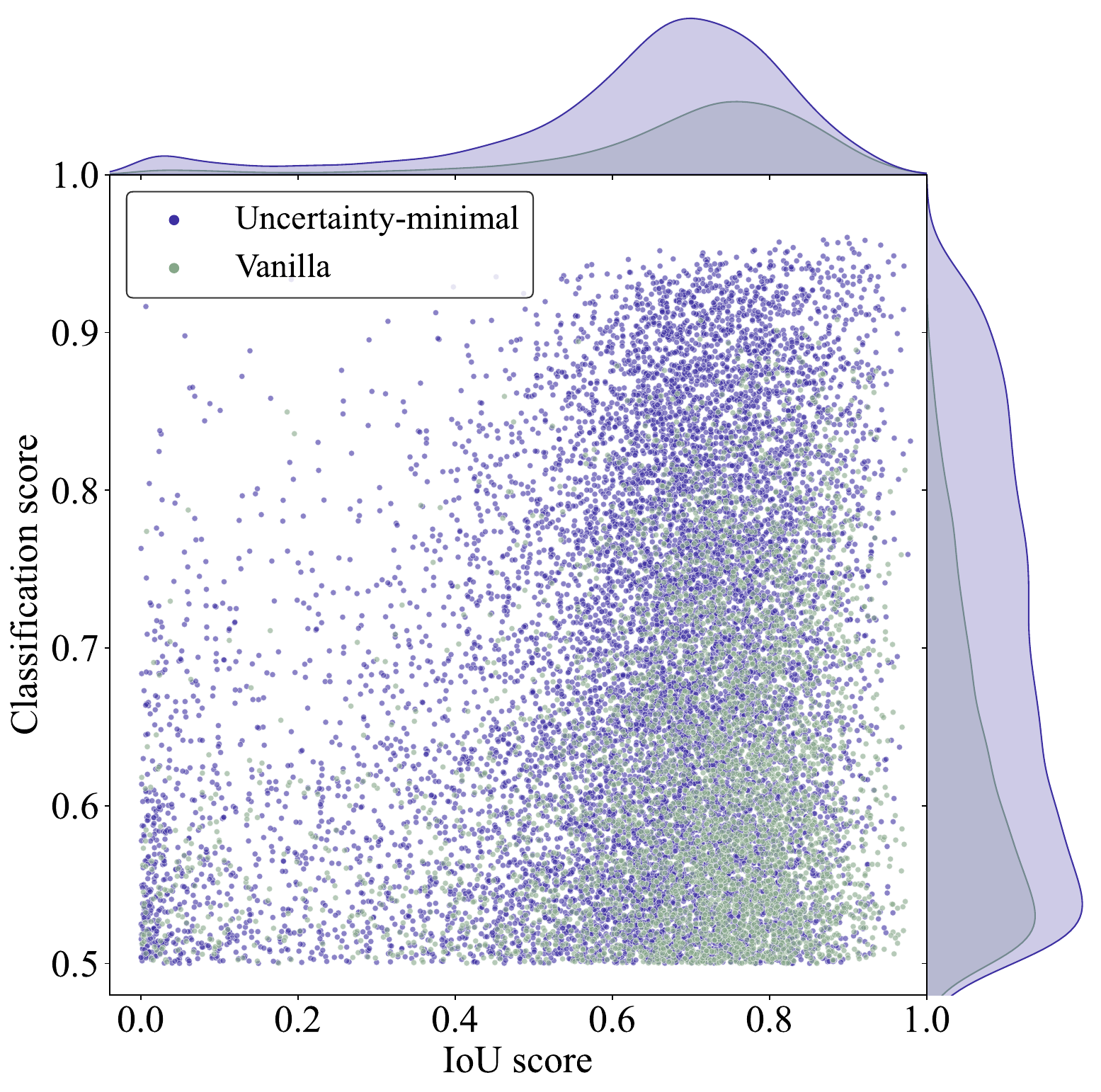}
\vspace*{-3mm}
\caption{Classification and IoU scores of the selected encoder features. \textcolor[rgb]{0.239, 0.188, 0.635}{Purple} and \textcolor[rgb]{0.525, 0.655, 0.537}{Green} dots represent the selected features from model trained with uncertainty-minimal query selection and vanilla query selection, respectively.}
\vspace*{-6mm}
\label{fig:iou-aware}
\end{figure}

%
\noindent \textbf{Effectiveness analysis.} To analyze the effectiveness of the uncertainty-minimal query selection, we visualize the classification scores and IoU scores of the selected features on COCO \texttt{val2017}, \Cref{fig:iou-aware}.
We draw the scatterplot with classification scores greater than $0.5$.
The purple and green dots represent the selected features from the model trained with uncertainty-minimal query selection and vanilla query selection, respectively.
The closer the dot is to the top right of the figure, the higher the quality of the corresponding feature, \ie, the more likely the predicted category and box are to describe the true object.
The top and right density curves reflect the number of dots for two types.

The most striking feature of the scatterplot is that the purple dots are concentrated in the top right of the figure, while the green dots are concentrated in the bottom right.
This shows that uncertainty-minimal query selection produces more high-quality encoder features. 
Furthermore, we perform quantitative analysis on two query selection schemes. 
There are $138\%$ more purple dots than green dots, \ie, more green dots with a classification score less than or equal to $0.5$, which can be considered low-quality features.
And there are $120\%$ more purple dots than green dots with both scores greater than $0.5$.
The same conclusion can be drawn from the density curves, where the gap between purple and green is most evident in the top right of the figure.
Quantitative results further demonstrate that the uncertainty-minimal query selection provides more features with accurate classification and precise location for queries, thereby improving the accuracy of the detector~(\textit{cf.} \cref{subsec:ablation iou-aware}).

\subsection{Scaled RT-DETR}
\label{subsec:scaled rt-detr}
Since real-time detectors typically provide models at different scales to accommodate different scenarios, RT-DETR also supports flexible scaling.
Specifically, for the hybrid encoder, we control the width by adjusting the embedding dimension and the number of channels, and the depth by adjusting the number of Transformer layers and \textit{RepBlock}s.
The width and depth of the decoder can be controlled by manipulating the number of object queries and decoder layers.
Furthermore, the speed of RT-DETR supports flexible adjustment by adjusting the number of decoder layers.
We observe that removing a few decoder layers at the end has minimal effect on accuracy, but greatly enhances inference speed~(\textit{cf.} \cref{subsec:ablation decoder}).
We compare the RT-DETR equipped with ResNet50 and ResNet101~\cite{he2016deep,he2019bag} to the $L$ and $X$ models of YOLO detectors.
Lighter RT-DETRs can be designed by applying other smaller~(\eg, ResNet18/34) or scalable~(\eg, CSPResNet~\cite{xu2022pp}) backbones with scaled encoder and decoder.
We compare the scaled RT-DETRs with the lighter~($S$ and $M$) YOLO detectors in Appendix, which outperform all $S$ and $M$ models in both speed and accuracy.
%
\section{Experiments}
\subsection{Comparison with SOTA}
\begin{table*}[!ht]
\footnotesize
\centering
\renewcommand{\arraystretch}{1.05}
\setlength{\tabcolsep}{3.85pt}
\begin{tabular*}{\textwidth}{ll | cccc | cccccc}
    \toprule
    \textbf{Model} &  \textbf{Backbone} & \textbf{\#Epochs} & \textbf{\#Params~(M)} & \textbf{GFLOPs} &  \textbf{FPS$_{bs=1}$} & \textbf{AP$^{val}$} & \textbf{AP$^{val}_{50}$} & \textbf{AP$^{val}_{75}$} & \textbf{AP$^{val}_S$} & \textbf{AP$^{val}_M$} & \textbf{AP$^{val}_L$} \\ 
    \midrule
    \hline
    \rowcolor[gray]{0.9}
    \multicolumn{12}{l}{\emph{Real-time Object Detectors}} \\
    YOLOv5-L~\cite{yolov5v7.0} & - & 300 & 46 & 109 & 54 & 49.0 & 67.3 & - & - & - & - \\
    YOLOv5-X~\cite{yolov5v7.0} & - & 300 & 86 & 205 & 43 & 50.7 & 68.9 & - & - & - & - \\
    PPYOLOE-L~\cite{xu2022pp} & - & 300 & 52 & 110  & 94 & 51.4 & 68.9 & 55.6 & 31.4 & 55.3 & 66.1 \\
    PPYOLOE-X~\cite{xu2022pp} & - & 300 & 98 & 206  & 60 & 52.3 & 69.9 & 56.5 & 33.3 & 56.3 & 66.4\\
    YOLOv6-L~\cite{li2023yolov6v3} & - & 300 & 59 & 150  & 99 & 52.8 & 70.3 & 57.7 & 34.4 & 58.1 & 70.1 \\
    YOLOv7-L~\cite{wang2023yolov7} & - &300 & 36 & 104  & 55 & 51.2 & 69.7 & 55.5 & 35.2 & 55.9 & 66.7 \\
    YOLOv7-X~\cite{wang2023yolov7} & - &300 & 71  & 189 & 45 & 52.9 & 71.1 & 57.4 & 36.9 & 57.7 & 68.6\\
    YOLOv8-L~\cite{yolov8} & - & - & 43 & 165  & 71 & 52.9 & 69.8 & 57.5 & 35.3 & 58.3 & 69.8 \\
    YOLOv8-X~\cite{yolov8} & - & - & 68 & 257  & 50 & 53.9 & 71.0 & 58.7 & 35.7 & 59.3 & 70.7 \\
    \midrule
    \hline
    \rowcolor[gray]{0.9}
    \multicolumn{12}{l}{\emph{End-to-end Object Detectors}} \\

    DETR-DC5~\cite{carion2020end} & R50 & 500 & 41 &  187 & - & 43.3 & 63.1 & 45.9 & 22.5 & 47.3 & 61.1 \\
    DETR-DC5~\cite{carion2020end} & R101 & 500 & 60 &  253 & - & 44.9 & 64.7 & 47.7 & 23.7 & 49.5 & 62.3 \\
    
    Anchor-DETR-DC5~\cite{wang2022anchor} & R50 & 50 & 39 &  172 & - & 44.2 & 64.7 & 47.5 & 24.7 & 48.2 & 60.6 \\
    Anchor-DETR-DC5~\cite{wang2022anchor} & R101 & 50 & - & - & - & 45.1 & 65.7 & 48.8 & 25.8 & 49.4 & 61.6 \\
    
    Conditional-DETR-DC5~\cite{meng2021conditional} & R50 & 108 & 44 & 195 & - & 45.1 & 65.4 & 48.5 & 25.3 & 49.0 & 62.2 \\
    Conditional-DETR-DC5~\cite{meng2021conditional} & R101 & 108 & 63 & 262 & - & 45.9 & 66.8 & 49.5 & 27.2 & 50.3 & 63.3 \\

    Efficient-DETR~\cite{yao2021efficient} & R50 & 36 & 35 &  210 & - & 45.1 & 63.1 & 49.1 & 28.3 & 48.4 & 59.0 \\
    Efficient-DETR~\cite{yao2021efficient} & R101 & 36 & 54 &  289 & - & 45.7 & 64.1 & 49.5 & 28.2 & 49.1 & 60.2 \\

    SMCA-DETR~\cite{gao2021fast} & R50 & 108 & 40 &  152  & - & 45.6 & 65.5 & 49.1 & 25.9 & 49.3 & 62.6 \\
    SMCA-DETR~\cite{gao2021fast} & R101 & 108 & 58 &  218  & - & 46.3 & 66.6 & 50.2 & 27.2 & 50.5 & 63.2 \\

    Deformable-DETR~\cite{zhu2020deformable} & R50 & 50 & 40 & 173 & - & 46.2 & 65.2 & 50.0 & 28.8 & 49.2 & 61.7 \\
    DAB-Deformable-DETR~\cite{liu2021dab} & R50 & 50 & 48 &  195 & - & 46.9 & 66.0 & 50.8 & 30.1 & 50.4 & 62.5 \\
    DAB-Deformable-DETR++~\cite{liu2021dab} & R50 & 50 & 47 & - & - & 48.7 & 67.2 & 53.0 & 31.4 & 51.6 & 63.9 \\
    DN-Deformable-DETR~\cite{li2022dn} & R50 & 50 & 48 &  195 & - & 48.6 & 67.4 & 52.7 & 31.0 & 52.0 & 63.7 \\
    DN-Deformable-DETR++~\cite{li2022dn} & R50 & 50 & 47 & - & - & 49.5 & 67.6 & 53.8 & 31.3 & 52.6 & 65.4 \\
    
    DINO-Deformable-DETR~\cite{zhang2022dino} & R50 & 36 & 47 &  279 & 5 & 50.9 & 69.0 & 55.3 & 34.6 & 54.1 & 64.6 \\

    \midrule
    \hline
    \rowcolor{aliceblue}
    \multicolumn{12}{l}{\emph{Real-time End-to-end Object Detector~(ours)}} \\
    
    RT-DETR & R50 & 72 & 42 & 136 & \textbf{108} & \textbf{53.1} & \textbf{71.3} & \textbf{57.7} & 34.8 & 58.0 & 70.0 \\ 
    RT-DETR & R101 & 72 & 76 & 259 & \textbf{74} & \textbf{54.3} & \textbf{72.7} & 58.6 & 36.0 & 58.8 & \textbf{72.1} \\
    \bottomrule
\end{tabular*}
\vspace*{-2mm}
\caption{
Comparison with SOTA~(only $L$ and $X$ models of YOLO detectors, see Appendix for the comparison with $S$ and $M$ models). We do not test the speed of other DETRs, except for DINO-Deformable-DETR~\cite{zhang2022dino} for comparison, as they are not real-time detectors. Our RT-DETR outperforms the state-of-the-art YOLO detectors and DETRs in both speed and accuracy.
}
\vspace*{-6mm}
\label{tab:main}
\end{table*}

%
\Cref{tab:main} compares RT-DETR with current real-time~(YOLOs) and end-to-end~(DETRs) detectors, where only the $L$ and $X$ models of the YOLO detector are compared, and the $S$ and $M$ models are compared in Appendix.
Our RT-DETR and YOLO detectors share a common input size of (640, 640), and other DETRs use an input size of (800, 1333).
The FPS is reported on T4 GPU with TensorRT FP16, and for YOLO detectors using official pre-trained models according to the end-to-end speed benchmark proposed in~\cref{subsec: benchmark}.
Our RT-DETR-R50 achieves $53.1\%$ AP and $108$ FPS, while RT-DETR-R101 achieves $54.3\%$ AP and $74$ FPS, outperforming state-of-the-art YOLO detectors of similar scale and DETRs with the same backbone in both speed and accuracy.
The experimental settings are shown in Appendix.

\noindent \textbf{Comparison with real-time detectors.}
We compare the end-to-end speed~(\textit{cf.} \cref{subsec: benchmark}) and accuracy of RT-DETR with YOLO detectors.
We compare RT-DETR with YOLOv5~\cite{yolov5v7.0}, PP-YOLOE~\cite{xu2022pp}, YOLOv6v3.0~\cite{li2023yolov6v3}~(hereinafter referred to as YOLOv6), YOLOv7~\cite{wang2023yolov7} and YOLOv8~\cite{yolov8}. 
Compared to YOLOv5-L / PP-YOLOE-L / YOLOv6-L, RT-DETR-R50 improves accuracy by $4.1\%$ / $1.7\%$ / $0.3\%$ AP, increases FPS by $100.0\%$ / $14.9\%$ / $9.1\%$, and reduces the number of parameters by $8.7\%$ / $19.2\%$ / $28.8\%$.
Compared to YOLOv5-X / PP-YOLOE-X, RT-DETR-R101 improves accuracy by $3.6\%$ / $2.0\%$, increases FPS by $72.1\%$ / $23.3\%$, and reduces the number of parameters by $11.6\%$ / $22.4\%$.
Compared to YOLOv7-L / YOLOv8-L, RT-DETR-R50 improves accuracy by $1.9\%$ / $0.2\%$ AP and increases FPS by $96.4\%$ / $52.1\%$.
Compared to YOLOv7-X / YOLOv8-X, RT-DETR-R101 improves accuracy by $1.4\%$ / $0.4\%$ AP and increases FPS by $64.4\%$ / $48.0\%$.
This shows that our RT-DETR achieves state-of-the-art real-time detection performance.

\noindent \textbf{Comparison with end-to-end detectors.}
We also compare RT-DETR with existing DETRs using the same backbone.
We test the speed of DINO-Deformable-DETR~\cite{zhang2022dino} according to the settings of the corresponding accuracy taken on COCO \texttt{val2017} for comparison, \ie, the speed is tested with TensorRT FP16 and the input size is (800, 1333). 
\Cref{tab:main} shows that RT-DETR outperforms all DETRs with the same backbone in both speed and accuracy.
Compared to DINO-Deformable-DETR-R50, RT-DETR-R50 improves the accuracy by $2.2\%$ AP and the speed by $21$ times~($108$ FPS vs $5$ FPS), both of which are significantly improved.

\subsection {Ablation Study on Hybrid Encoder}
\label{subsec:ablation encoder}
%
We evaluate the indicators of the variants designed in~\cref{subsec:encoder}, including AP~(trained with $1\times$ configuration), the number of parameters, and the latency, \Cref{tab:encoder}.
Compared to baseline A, variant B improves accuracy by $1.9\%$ AP and increases the latency by $54\%$.
This proves that the intra-scale feature interaction is significant, but the single-scale Transformer encoder is computationally expensive.
Variant C delivers a $0.7\%$ AP improvement over B and increases the latency by $20\%$.
This shows that the cross-scale feature fusion is also necessary but the multi-scale Transformer encoder requires higher computational cost.
Variant D delivers a $0.8\%$ AP improvement over C, but reduces latency by $8\%$, suggesting that decoupling intra-scale interaction and cross-scale fusion not only reduces computational cost but also improves accuracy.
Compared to variant D, $D_{\mathbfcal{S}_5}$ reduces the latency by $35\%$ but delivers $0.4\%$ AP improvement, demonstrating that intra-scale interactions of lower-level features are not required.
Finally, variant E delivers $1.5\%$ AP improvement over D.
Despite a $20\%$ increase in the number of parameters, the latency is reduced by $24\%$, making the encoder more efficient.
This shows that our hybrid encoder achieves a better trade-off between speed and accuracy.
\begin{table}[!t]
\centering
\renewcommand{\arraystretch}{1.0}
\setlength{\tabcolsep}{14.45pt}
\begin{tabular*}{\linewidth}{cccc}
  \toprule
   \textbf{Variant} & \textbf{\makecell[c]{AP\\\small(\%)}} & \textbf{\makecell[c]{\#Params\\\small(M)}} & \textbf{\makecell[c]{Latency\\\small(ms)}} \\ 
  \midrule
  \midrule
  A & 43.0 & 31 & 7.2 \\
  B & 44.9 & 32 & 11.1 \\
  C & 45.6 & 32 & 13.3 \\
  D & 46.4 & 35 & 12.2 \\
  D$_{\mathbfcal{S}_5}$ & 46.8 & 35 & 7.9 \\
  \rowcolor{aliceblue}
  E & 47.9 & 42 & 9.3 \\
  \bottomrule
\end{tabular*}
\vspace*{-2mm}
\caption{The indicators of the set of variants illustrated in~\Cref{fig:variants}.}
\label{tab:encoder}
\vspace*{-2mm}
\end{table}

\subsection {Ablation Study on Query Selection}
\label{subsec:ablation iou-aware}

We conduct an ablation study on uncertainty-minimal query selection, and the results are reported on RT-DETR-R50 with $1\times$ configuration, \Cref{tab:iou-aware}.
The query selection in RT-DETR selects the top $K~(K=300)$ encoder features according to the classification scores as the content queries, and the prediction boxes corresponding to the selected features are used as initial position queries.
We compare the encoder features selected by the two query selection schemes on COCO \texttt{val2017} and calculate the proportions of classification scores greater than $0.5$ and both classification and IoU scores greater than $0.5$, respectively.
The results show that the encoder features selected by uncertainty-minimal query selection not only increase the proportion of high classification scores~($0.82\%$ vs $0.35\%$) but also provide more high-quality features~($0.67\%$ vs $0.30\%$).
We also evaluate the accuracy of the detectors trained with the two query selection schemes on COCO \texttt{val2017}, where the uncertainty-minimal query selection achieves an improvement of $0.8\%$ AP~($48.7\%$ AP vs $47.9\%$ AP).
\begin{table}[!t]
\centering
\renewcommand{\arraystretch}{1}
\setlength{\tabcolsep}{6.4pt}
\begin{tabular*}{\linewidth}{cccc}
  \toprule 
  \textbf{Query selection} & \textbf{\makecell[c]{AP\\\small(\%)}} & \textbf{\makecell[c]{Prop$_{cls}$$ \uparrow $\\\small(\%)}} & \textbf{\makecell[c]{Prop$_{both}$$ \uparrow $\\\small(\%)}} \\
  \midrule
  \midrule
  \textbf{Vanilla} & 47.9 & 0.35 & 0.30 \\
  \rowcolor{aliceblue}
  \textbf{Uncertainty-minimal} & 48.7 & 0.82 & 0.67 \\
  \bottomrule
\end{tabular*}
\vspace*{-2mm}
\caption{Results of the ablation study on uncertainty-minimal query selection. \textbf{Prop$_{cls}$} and \textbf{Prop$_{both}$} represent the proportion of classification score and both scores greater than $0.5$ respectively.}
\vspace*{-6mm}
\label{tab:iou-aware}
\end{table}
\subsection {Ablation Study on Decoder}
\label{subsec:ablation decoder}
%
\Cref{tab:decoder} shows the inference latency and accuracy of each decoder layer of RT-DETR-R50 trained with different numbers of decoder layers.
When the number of decoder layers is set to $6$, the RT-DETR-R50 achieves the best accuracy $53.1\%$ AP.
%
%
Furthermore, we observe that the difference in accuracy between adjacent decoder layers gradually decreases as the index of the decoder layer increases.
Taking the column RT-DETR-R50-Det$^6$ as an example, using $5$-th decoder layer for inference only loses $0.1\%$ AP~($53.1\%$ AP vs $53.0\%$ AP) in accuracy, while reducing latency by $0.5$ ms~($9.3$ ms vs $8.8$ ms).
Therefore, RT-DETR supports flexible speed tuning by adjusting the number of decoder layers without retraining, thus improving its practicality.
\begin{table}[!t]
\centering
\renewcommand{\arraystretch}{1.0}
\setlength{\tabcolsep}{9.45pt}
\begin{tabular*}{\linewidth}{c | cccc | c}
  \toprule
  \multirow{2}*{\textbf{ID}} & \multicolumn{4}{c|}{\textbf{AP\small(\%)}} &\multirow{2}*{\textbf{\makecell[c]{Latency\\\small(ms)}}}\\ 
  \cline{2-5}
  & \textbf{Det$^4$} & \textbf{Det$^5$} & \textbf{Det$^6$} & \textbf{Det$^7$} & \\
  \midrule
  \midrule
  7 & - & - & - & 52.6 & 9.6\\
  6 & - & - & \cellcolor{aliceblue}\textbf{53.1} & 52.6 & 9.3 \\
  5 & - & 52.9 & 53.0 & 52.5 & 8.8\\
  4 & 52.7 & 52.7 & 52.7 & 52.1 & 8.3\\
  3 & 52.4 & 52.3 & 52.4 & 51.5 & 7.9\\
  2 & 51.6 & 51.3 & 51.3 & 50.6 & 7.5\\
  1 & 49.6 & 48.8 & 49.1 & 48.3 & 7.0\\
  \bottomrule
\end{tabular*}
\caption{Results of the ablation study on decoder. \textbf{ID} indicates decoder layer index. \textbf{Det$^k$} represents detector with $k$ decoder layers. All results are reported on RT-DETR-R50 with $6\times$ configuration.}
\vspace*{-5mm}
\label{tab:decoder}
\end{table}
%


\section{Limitation and Discussion}
\label{sec:limit}
\noindent \textbf{Limitation.} 
Although the proposed RT-DETR outperforms the state-of-the-art real-time detectors and end-to-end detectors with similar size in both speed and accuracy, it shares the same limitation as the other DETRs, \ie, the performance on small objects is still inferior than the strong real-time detectors.
According to~\Cref{tab:main}, RT-DETR-R50 is $0.5\%$ AP lower than the highest AP$_S^{val}$ in the $L$ model~(YOLOv8-L) and RT-DETR-R101 is $0.9\%$ AP lower than the highest AP$_S^{val}$ in the $X$ model~(YOLOv7-X).
We hope that this problem will be addressed in future work.

\noindent \textbf{Discussion.}
Existing large DETR models~\cite{chen2022groupv2,zhang2022dino,zong2023detrs,cai2022reversible,yang2022focal,ren2023strong} have demonstrated impressive performance on COCO \texttt{test-dev}~\cite{lin2014microsoft} leaderboard. 
The proposed RT-DETR at different scales preserves decoders homogeneous to other DETRs, which makes it possible to distill our lightweight detector with high accuracy pre-trained large DETR models.
We believe that this is one of the advantages of RT-DETR over other real-time detectors and could be an interesting direction for future exploration.
%
\section{Conclusion}
\label{sec:conclusion}
In this work, we propose a real-time end-to-end detector, called RT-DETR, which successfully extends DETR to the real-time detection scenario and achieves state-of-the-art performance.
RT-DETR includes two key enhancements: an efficient hybrid encoder that expeditiously processes multi-scale features, and the uncertainty-minimal query selection that improves the quality of initial object queries.
%
Furthermore, RT-DETR supports flexible speed tuning without retraining and eliminates the inconvenience caused by two NMS thresholds, facilitating its practical application.
RT-DETR, along with its model scaling strategy, broadens the technical approach to real-time object detection, offering new possibilities beyond YOLO for diverse real-time scenarios.
%
We hope that RT-DETR can be put into practice.

\smallskip
\noindent \textbf{Acknowledgements.} This work was supported in part by the National Key R\&D Program of China (No. 2022ZD0118201), Natural Science Foundation of China (No. 61972217, 32071459, 62176249, 62006133, 62271465), and the Shenzhen Medical Research Funds in China (No. B2302037).
Thanks to Chang Liu, Zhennan Wang and Kehan Li for helpful suggestions on writing and presentation.

{
    \small
    \bibliographystyle{ieeenat_fullname}
    \bibliography{main}
}

\clearpage
\setcounter{page}{1}
\maketitleappendix
\renewcommand{\thefootnote}{\fnsymbol{footnote}}

\renewcommand{\thetable}{\Alph{table}}
\renewcommand{\theequation}{\Alph{equation}}
\renewcommand{\thefigure}{\Alph{figure}}

\setcounter{table}{0}
\setcounter{section}{0}
\setcounter{figure}{0}
\setcounter{equation}{0}

\definecolor{darkgreen}{HTML}{539165}
\newcommand{\increase}[1]{{
  \fontsize{9.5pt}{0.5em}\selectfont({\color{darkgreen}{$\uparrow$~\textbf{#1}}})
}}
\section{Experimental Settings}

\noindent \textbf{Dataset and metric.}
We conduct experiments on COCO~\cite{lin2014microsoft} and Objects365~\cite{shao2019objects365}, where RT-DETR is trained on COCO \texttt{train2017} and validated on COCO \texttt{val2017} dataset.
We report the standard COCO metrics, including AP~(averaged over uniformly sampled IoU thresholds ranging from 0.50-0.95 with a step size of 0.05), AP$_{50}$, AP$_{75}$, as well as AP at different scales: AP$_S$, AP$_M$, AP$_L$.

\noindent \textbf{Implementation details.}
We use ResNet~\cite{he2016deep,he2019bag} pretrained on ImageNet~\cite{russakovsky2015imagenet,DBLP:journals/corr/abs-2103-05959} as the backbone and the learning rate strategy of the backbone follows~\cite{carion2020end}.
In the hybrid encoder, AIFI consists of $1$ Transformer layer and the fusion block in CCFF consists of $3$ \textit{RepBlock}s.
We leverage the uncertainty-minimal query selection to select top $300$ encoder features to initialize object queries of the decoder.
The training strategy and hyperparameters of the decoder almost follow DINO~\cite{zhang2022dino}.
We train RT-DETR with the AdamW~\cite{loshchilov2018decoupled} optimizer using four NVIDIA Tesla V100 GPUs with a batch size of 16 and apply the exponential moving average~(EMA) with $ema\_decay=0.9999$.
The $1\times$ configuration means that the total epoch is $12$, and the final reported results adopt the $6\times$ configuration.
The data augmentation applied during training includes \textit{random\_\{color distort, expand, crop, flip, resize\}} operations, following~\cite{xu2022pp}.
The main hyperparameters of RT-DETR are listed in~\Cref{tab:hyper}~(refer to RT-DETR-R50 for detailed configuration).

\begin{table}[!ht]
\centering
\renewcommand{\arraystretch}{1.15}
\setlength{\tabcolsep}{15pt}
\begin{tabular}{l|l}
  \toprule
  \textbf{Item} & \textbf{Value} \\
  \midrule
  \midrule
  optimizer & AdamW \\
  base learning rate & 1e-4  \\
  learning rate of backbone & 1e-5 \\
  freezing BN & True \\
  linear warm-up start factor & 0.001 \\
  linear warm-up steps & 2000 \\ 
  weight decay & 0.0001 \\
  clip gradient norm & 0.1 \\
  ema decay & 0.9999 \\
  number of AIFI layers & 1 \\
  number of \textit{RepBlock}s & 3 \\
  embedding dim & 256 \\
  feedforward dim & 1024 \\
  nheads & 8 \\
  number of feature scales & 3 \\
  number of decoder layers & 6 \\
  number of queries & 300 \\
  decoder npoints & 4 \\
  class cost weight & 2.0 \\
  $\alpha$ in class cost & 0.25 \\
  $\gamma$ in class cost & 2.0 \\
  bbox cost weight & 5.0 \\
  GIoU cost weight & 2.0 \\
  class loss weight & 1.0 \\
  $\alpha$ in class loss & 0.75 \\
  $\gamma$ in class loss & 2.0 \\
  bbox loss weight & 5.0 \\
  GIoU loss weight & 2.0 \\
  denoising number & 200 \\
  label noise ratio & 0.5 \\
  box noise scale & 1.0 \\
  \bottomrule
\end{tabular}
\caption{Main hyperparameters of RT-DETR.}
\vspace*{-6mm}
\label{tab:hyper}
\end{table}

\section{Comparison with Lighter YOLO Detectors}
\label{sec: comparison}

\begin{table*}[!ht]
\centering
\renewcommand{\arraystretch}{1.15}
\resizebox{\linewidth}{!}{
    \begin{tabular}{l | cccc | cccccc}
        \toprule
        \textbf{Model} & \textbf{\#Epochs} & \textbf{\#Params~(M)} & \textbf{GFLOPs} &  \textbf{FPS$_{bs=1}$} & \textbf{AP$^{val}$} & \textbf{AP$^{val}_{50}$} & \textbf{AP$^{val}_{75}$} & \textbf{AP$^{val}_S$} & \textbf{AP$^{val}_M$} & \textbf{AP$^{val}_L$} \\ 
        \midrule
        \hline
        \rowcolor[gray]{0.9}
        \multicolumn{11}{l}{\emph{$S$ and $M$ models of YOLO Detectors}} \\
        YOLOv5-S\cite{yolov5v7.0}  & 300 & 7.2 & 16.5 & 74 & 37.4 & 56.8 & - & - & - & - \\
        YOLOv5-M\cite{yolov5v7.0}  & 300 & 21.2 & 49.0 & 64 & 45.4 & 64.1 & - & - & - & - \\
        PPYOLOE-S\cite{xu2022pp}  & 300 & 7.9 & 17.4  & 218 & 43.0 & 59.6 & 47.1 & 25.9 & 47.4 & 58.6 \\
        PPYOLOE-M\cite{xu2022pp}  & 300 & 23.4 & 49.9  & 131 & 48.9 & 65.8 & 53.7 & 30.8 & 53.4 & 65.3\\
        YOLOv6-S\cite{li2023yolov6v3} & 300 & 18.5 & 45.3  & 201 & 45.0 & 61.8 & 48.9 & 24.3 & 50.2 & 62.7 \\
        YOLOv6-M\cite{li2023yolov6v3} & 300 & 34.9 & 85.8  & 121 & 50.0 & 66.9 & 54.6 & 30.6 & 55.4 & 67.3 \\
        YOLOv8-S\cite{yolov8} & - & 11.2 & 28.6  & 136 & 44.9 & 61.8 & 48.6 & 25.7 & 49.9 & 61.0 \\
        YOLOv8-M\cite{yolov8} & - & 25.9 & 78.9  & 97 & 50.2 & 67.2 & 54.6 & 32.0 & 55.7 & 66.4 \\
        \midrule
        \hline
        \rowcolor[gray]{0.9}
        \multicolumn{11}{l}{\emph{Scaled RT-DETRs}} \\
        Scaled RT-DETR-R50-Dec$^2$ & 72 & 36$^\dagger$ & 98.4 & \textbf{154} & \textbf{50.3} & \textbf{68.4} & 54.5 & \textbf{32.2} & 55.2 & \textbf{67.5} \\
        Scaled RT-DETR-R50-Dec$^3$ & 72 & 36$^\dagger$ & 100.1 & \textbf{145} & \textbf{51.3} & \textbf{69.6} & \textbf{55.4} & \textbf{33.6} & \textbf{56.1} & \textbf{68.6} \\ 
        Scaled RT-DETR-R50-Dec$^4$ & 72 & 36$^\dagger$ & 101.8 & \textbf{137} & \textbf{51.8} & \textbf{70.0} & \textbf{55.9} & \textbf{33.7} & \textbf{56.4} & \textbf{69.4} \\
        \rowcolor{aliceblue!80}
        Scaled RT-DETR-R50-Dec$^5$ & 72 & 36$^\dagger$ & 103.5 & \textbf{132} & \textbf{52.1} & \textbf{70.5} & \textbf{56.2} & \textbf{34.3} & \textbf{56.9} & \textbf{69.9} \\ 
        Scaled RT-DETR-R50-Dec$^6$ & 72 & 36 & 105.2 & 125 & \textbf{52.2} & \textbf{70.6} & \textbf{56.4} & \textbf{34.4} & \textbf{57.0} & \textbf{70.0} \\ 
        \midrule
        Scaled RT-DETR-R34-Dec$^2$ & 72 & 31$^\dagger$ & 89.3 & 185 & \textbf{47.4} & \textbf{64.7} & \textbf{51.3} & \textbf{28.9 }& \textbf{51.0} & \textbf{64.2} \\
        Scaled RT-DETR-R34-Dec$^3$ & 72 & 31$^\dagger$ & 91.0 & 172 & \textbf{48.5} & \textbf{66.2} & \textbf{52.3} & \textbf{30.2} & \textbf{51.9} & \textbf{66.2} \\
        Scaled RT-DETR-R34-Dec$^4$ & 72 & 31 & 92.7 & 161 & \textbf{48.9} & \textbf{66.8} & \textbf{52.9} & \textbf{30.6} & \textbf{52.4} & \textbf{66.3} \\
        \midrule
        \rowcolor{aliceblue!80}
        Scaled RT-DETR-R18-Dec$^2$ & 72 & 20$^\dagger$ & 59.0 & \textbf{238} & \textbf{45.5} & \textbf{62.5} & \textbf{49.4} & \textbf{27.8} & 48.7 & 61.7 \\
        Scaled RT-DETR-R18-Dec$^3$ & 72 & 20 & 60.7 & 217 & \textbf{46.5} & \textbf{63.8} & \textbf{50.4} & \textbf{28.4} & 49.8 & \textbf{63.0} \\
        \bottomrule
    \end{tabular}
    } 
\caption{Comparison with $S$ and $M$ models of YOLO detectors. The FPS of YOLO detectors are reported on T4 GPU with TensorRT FP16 using official pre-trained models according to the proposed end-to-end speed benchmark. $\dagger$ denotes the number of parameters during the training, not inference.}
\label{tab:comparison}
\end{table*}
\begin{table*}[!ht]
\centering
\renewcommand{\arraystretch}{1.15}
\resizebox{0.95\linewidth}{!}{
   \begin{tabular}{l | cccc | cccccc}
   \toprule
   \textbf{Model} & \textbf{\#Epochs} & \textbf{\#Params~(M)} & \textbf{GFLOPs} &  \textbf{FPS$_{bs=1}$} & \textbf{AP$^{val}$} & \textbf{AP$^{val}_{50}$} & \textbf{AP$^{val}_{75}$} & \textbf{AP$^{val}_S$} & \textbf{AP$^{val}_M$} & \textbf{AP$^{val}_L$} \\ 
  \midrule
  \midrule
   RT-DETR-R18 & 60 & 20 & 61 & 217 & 49.2 \increase{2.7} & 66.6 & 53.5 & 33.2 & 52.3 & 64.8 \\
   RT-DETR-R50 & 24 & 42 & 136 &  108 & 55.3 \increase{2.2} & 73.4 & 60.1 & 37.9 & 59.9 & 71.8 \\ 
   RT-DETR-R101 & 24 & 76 & 259 & 74 & 56.2 \increase{1.9} & 74.6 & 61.3 & 38.3 & 60.5 & 73.5 \\
  \bottomrule
  \end{tabular}
  }
\caption{Fine-tuning results on COCO \texttt{val2017} with pre-training on Objects365.}
\vspace*{-2mm}
\label{tab:obj365_results}
\end{table*}
%
To adapt to diverse real-time detection scenarios, we develop lighter scaled RT-DETRs by scaling the encoder and decoder with ResNet50/34/18~\cite{he2016deep}.
Specifically, we halve the number of channels in the \textit{RepBlock}, while leaving other components unchanged, and obtain a set of RT-DETRs by adjusting the number of decoder layers during inference.
We compare the scaled RT-DETRs with the $S$ and $M$ models of YOLO detectors in~\Cref{tab:comparison}.
The number of decoder layers used by scaled RT-DETR-R50/34/18 during training is 6/4/3 respectively, and Dec$^k$ indicates that $k$ decoder layers are used during inference.
Our RT-DETR-R50-Dec$^{2-5}$ outperform all $M$ models of YOLO detectors in both speed and accuracy, while RT-DETR-R18-Dec$^2$ outperforms all $S$ models.
Compared to the state-of-the-art $M$ model~(YOLOv8-M~\cite{yolov8}), RT-DETR-R50-Dec$^5$ improves accuracy by $0.9\%$ AP and increases FPS by $36\%$.
Compared to the state-of-the-art $S$ model~(YOLOv6-S~\cite{li2023yolov6v3}), RT-DETR-R18-Dec$^2$ improves accuracy by $0.5\%$ AP and increases FPS by $18\%$.
This shows that RT-DETR is able to outperform the lighter YOLO detectors in both speed and accuracy by simple scaling.

\section{Large-scale Pre-training for RT-DETR}
\label{sec: obj365_results}
%
We pre-train RT-DETR on the larger Objects365\cite{shao2019objects365} dataset and then fine-tune it on COCO to achieve higher performance.
As shown in Table~\ref{tab:obj365_results}, we perform experiments on RT-DETR-R18/50/101 respectively. All three models are pre-trained on Objects365 for 12 epochs, and RT-DETR-R18 is fine-tuned on COCO for 60 epochs, while RT-DETR-R50 and RT-DETR-R101 are fine-tuned for 24 epochs.
Experimental results show that RT-DETR-R18/50/101 is improved by $2.7\%$/$2.2\%$/$1.9\%$ AP on COCO \texttt{val2017}. The surprising improvement further demonstrates the potential of RT-DETR and provides the strongest real-time object detector for various real-time scenarios in the industry.
%
%

\section{Visualization of Predictions with Different Post-processing Thresholds}
\label{sec: visualization}

\begin{figure*}[!ht]
\centering
\includegraphics[width=0.95\linewidth]{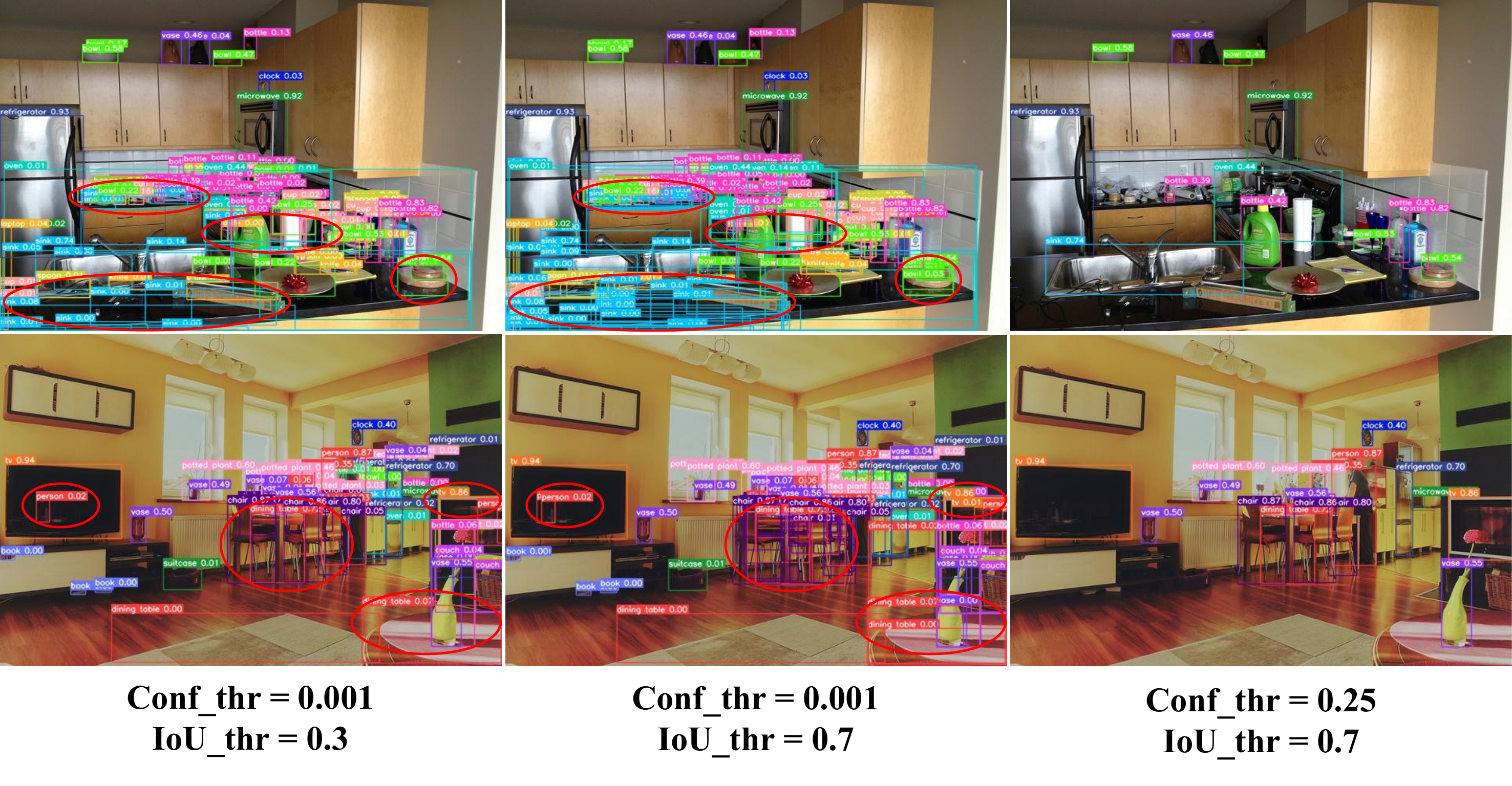}
\vspace*{-3mm}
\caption{Visualization of YOLOv8-L~\cite{yolov8} predictions with different NMS thresholds.}
\label{fig:vis_yolov8}
\end{figure*}
\begin{figure*}[!ht]
\centering
\includegraphics[width=0.95\linewidth]{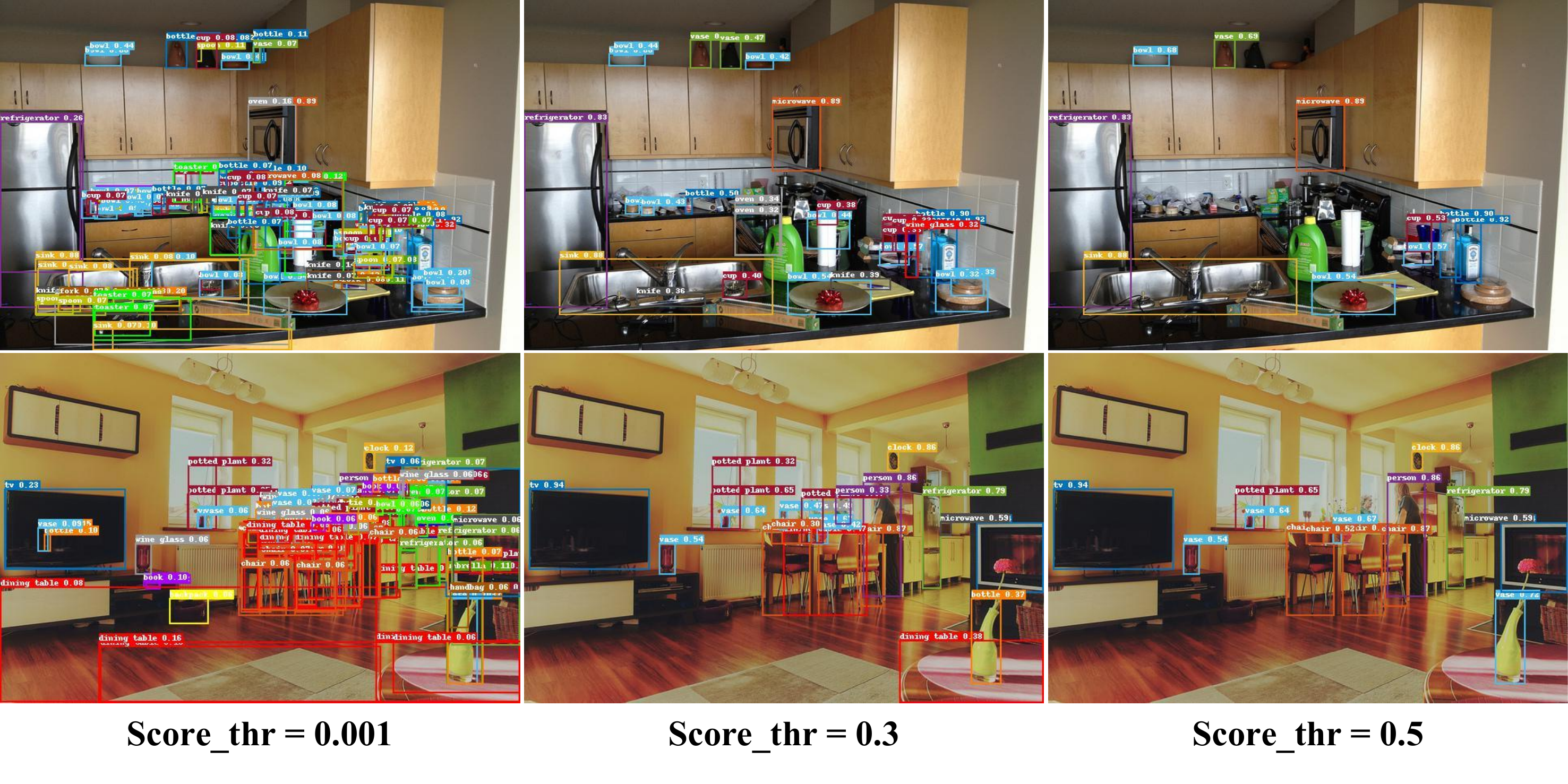}
\vspace*{-3mm}
\caption{Visualization of RT-DETR-R50 predictions with different score thresholds.}
\label{fig:vis_rtdetr}
\vspace*{-2mm}
\end{figure*}
%
To intuitively demonstrate the impact of post-processing on the detector, we visualize the predictions produced by YOLOv8~\cite{yolov8} and RT-DETR using different post-processing thresholds, as shown in \Cref{fig:vis_yolov8} and \Cref{fig:vis_rtdetr}, respectively.
We show the predictions for two randomly selected samples from COCO \texttt{val2017} by setting different NMS thresholds for YOLOv8-L and score thresholds for RT-DETR-R50.

There are two NMS thresholds: confidence threshold and IoU threshold, both of which affect the detection results.
The higher the confidence threshold, the more prediction boxes are filtered out and the number of false negatives increases.
However, using a lower confidence threshold, \eg, $0.001$, results in a large number of redundant boxes and increases the number of false positives.
The higher the IoU threshold, the fewer overlapping boxes are filtered out in each round of screening, and the number of false positives increases~(the position marked by the red circle in~\Cref{fig:vis_yolov8}).
Nevertheless, adopting a lower IoU threshold will result in true positives being deleted if there are overlapping or mutually occluding objects in the input.
The confidence threshold is relatively straightforward to process predicted boxes and therefore easy to set, whereas the IoU threshold is difficult to set accurately.
Considering that different scenarios place different emphasis on recall and accuracy, \eg, the general detection scenario requires the lower confidence threshold and the higher IoU threshold to increase the recall, while the dedicated detection scenario requires the higher confidence threshold and the lower IoU threshold to increase the accuracy, it is necessary to carefully select the appropriate NMS thresholds for different scenarios.

RT-DETR utilizes bipartite matching to predict the one-to-one object set, eliminating the need for suppressing overlapping boxes. 
Instead, it directly filters out low-confidence boxes with a score threshold.
Similar to the confidence threshold used in NMS, the score threshold can be adjusted in different scenarios based on the specific emphasis to achieve optimal detection performance.
Thus, setting the post-processing threshold in RT-DETR is straightforward and does not affect the inference speed, enhancing the adaptability of real-time detectors across various scenarios.

\section{Visualization of RT-DETR Predictions}
\label{sec: visualization_rtdetr}

\begin{figure*}[!ht]
\centering
\includegraphics[width=0.95\linewidth]{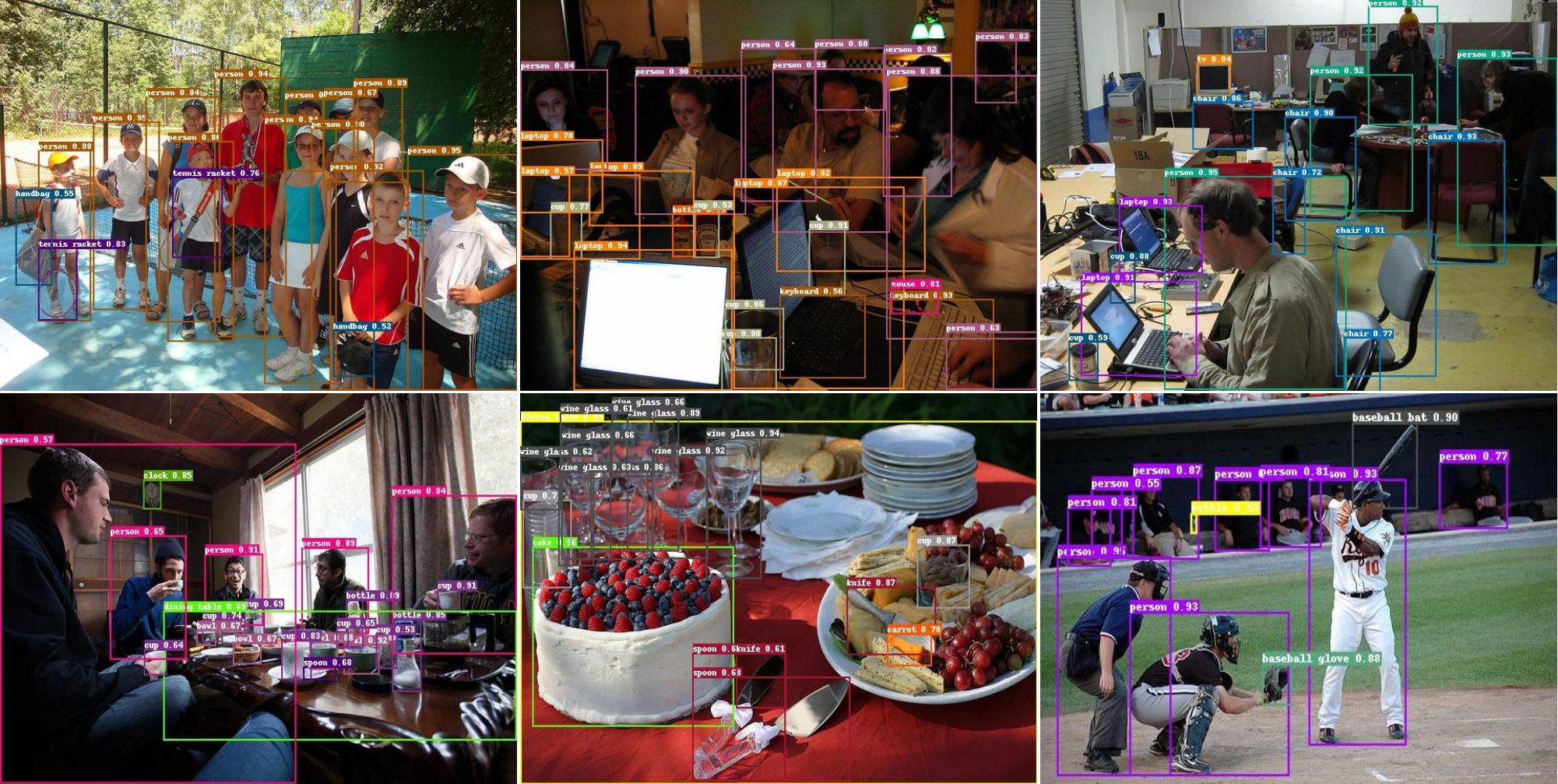}
\caption{Visualization of RT-DETR-R101 predictions in complex scenarios~(score threshold=0.5).}
\label{fig:vis_complex}
\end{figure*}

\begin{figure*}[!ht]
\centering
\includegraphics[width=0.95\linewidth]{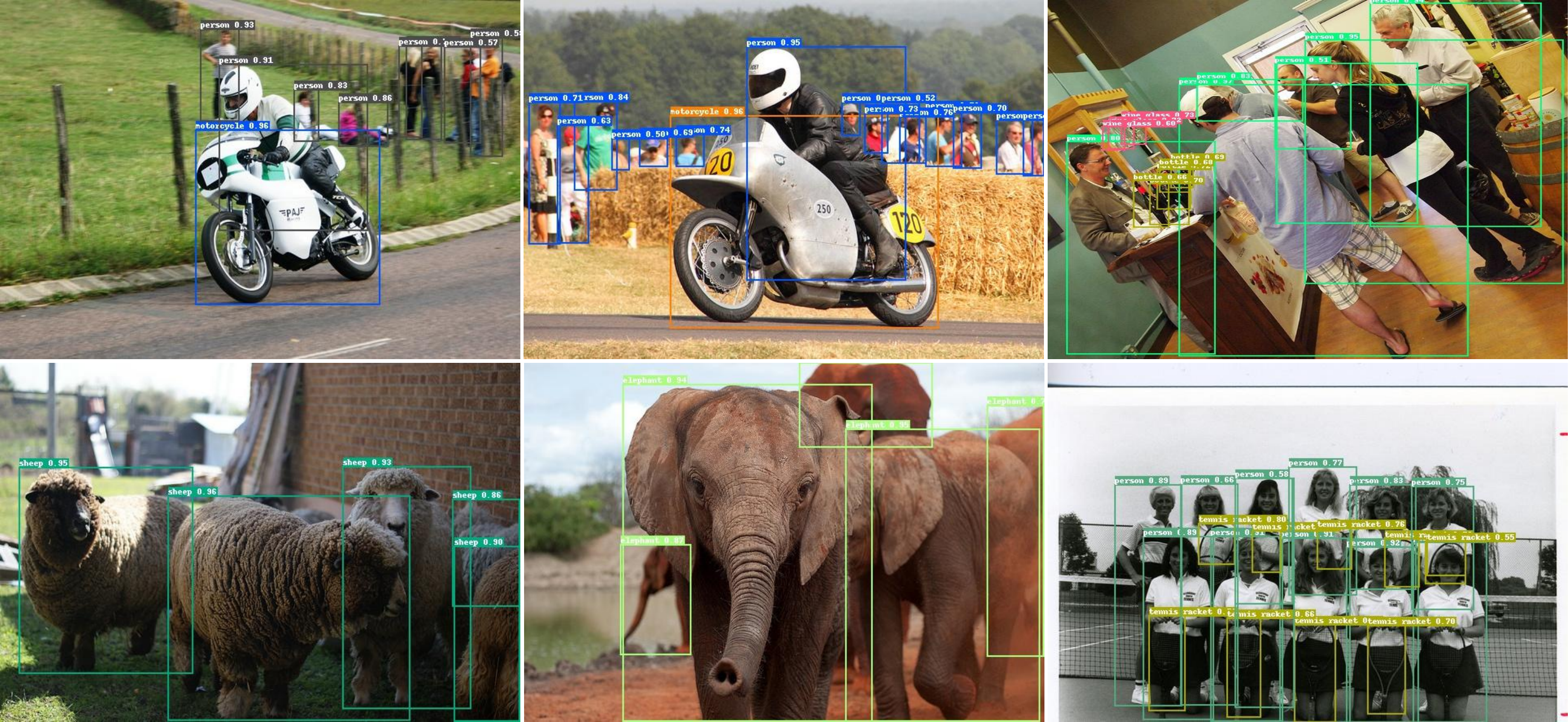}
\caption{Visualization of RT-DETR-R101 predictions under difficult conditions, including motion blur, rotation, and occlusion~(score threshold=0.5).}
\label{fig:vis_difficult}
\vspace*{-2mm}
\end{figure*}
%
We select several samples from the COCO \texttt{val2017} to showcase the detection performance of RT-DETR in complex scenarios and challenging conditions~(refer to \Cref{fig:vis_complex} and \Cref{fig:vis_difficult}). 
In complex scenarios, RT-DETR demonstrates its capability to detect diverse objects, even when they are small or densely packed, \eg, cups, wine glasses, and individuals. 
Moreover, RT-DETR successfully detects objects under various difficult conditions, including motion blur, rotation, and occlusion. 
These predictions substantiate the excellent detection performance of RT-DETR.
\end{document}